\documentclass[10pt,twocolumn,letterpaper]{article}

\usepackage{wacv}              % To produce the CAMERA-READY version
\usepackage{graphicx}
\usepackage{amsmath}
\usepackage{amssymb}
\usepackage{booktabs}

\usepackage[pagebackref,breaklinks,colorlinks]{hyperref}

\usepackage[capitalize]{cleveref}
\crefname{section}{Sec.}{Secs.}
\Crefname{section}{Section}{Sections}
\Crefname{table}{Table}{Tables}
\crefname{table}{Tab.}{Tabs.}

\def\wacvPaperID{*****} % *** Enter the WACV Paper ID here
\def\confName{WACV}
\def\confYear{2024}

\begin{document}

%%%%%%%%% TITLE - PLEASE UPDATE
\title{ReIDTracker\_Sea: the technical report of BoaTrack and  SeaDronesSee-MOT challenge at MaCVi}

\author{Kaer Huang\\
Lenovo Research\\
{\tt\small huangke1@lenovo.com}
% For a paper whose authors are all at the same institution,
% omit the following lines up until the closing ``}''.
% Additional authors and addresses can be added with ``\and'',
% just like the second author.
% To save space, use either the email address or home page, not both
\and
Aiguo Zheng\\
Lenovo\\
{\tt\small zhengag@lenovo.com}
\and
Weitu Chong\\
Fudan University\\
{\tt\small wtzhong22@m.fudan.edu.cn}
\and
Kanokphan Lertniphonphan\\
Lenovo Research\\
{\tt\small klertniphonp@lenovo.com}
\and
Jun Xie\\
Lenovo Research\\
{\tt\small xiejun@lenovo.com}
\and
Feng Chen\\
Lenovo Research\\
{\tt\small chenfeng13@lenovo.com}
\and
Jian Li\\
Lenovo\\
{\tt\small lijian30@lenovo.com}
\and
Zhepeng Wang\\
Lenovo Research\\
{\tt\small wangzpb@lenovo.com}
}
\maketitle

%%%%%%%%% ABSTRACT
\begin{abstract}
 Multi-Object Tracking is one of the
most important technologies in maritime computer vision.
Our solution tries to explore Multi-Object Tracking in maritime Unmanned Aerial vehicles (UAVs) and Unmanned Surface Vehicles (USVs) usage scenarios. Most of the current Multi-Object Tracking algorithms require complex association strategies and association information (2D location and motion, 3D motion, 3D depth, 2D appearance) to achieve better performance, which makes the entire tracking system extremely complex and heavy. At the same time, most of the current Multi-Object Tracking algorithms still require video annotation data which is costly to obtain for training. Our solution tries to explore Multi-Object Tracking in a completely unsupervised way. The scheme accomplishes instance representation learning by using self-supervision on ImageNet. Then, by cooperating with high-quality detectors, the multi-target tracking task can be completed simply and efficiently. The scheme achieved top 3 performance on both
UAV-based Multi-Object Tracking with Reidentification and USV-based Multi-Object Tracking benchmarks and the solution won the championship in many multiple Multi-Object Tracking competitions. such as BDD100K MOT,MOTS, Waymo 2D MOT.

\end{abstract}

%%%%%%%%% BODY TEXT
\section{Introduction}
\label{sec:intro}

Our solution considers whether we can achieve SOTA only based on high-performance detection and appearance models.  We use CBNetV2 Swin-B \cite{liang2021cbnetv2,huang2023reidtrack,huang1st} as the detection model and self-supervised learning MoCo-v2 \cite{he2020momentum} as a high-quality appearance model. We removed all motion information, including the Kalman filter and IoU mapping. We also introduce ByteTrack \cite{zhang2021bytetrack,kiefer20231st,kristan2023first} innovation to associate the low-score detection boxes and the high-score ones. The top-down view for UAV-Based datasets resulted in a low overlap of objects, we lowered the NMS threshold to adapt it.

\begin{figure*}[h!]
    \centering
    \includegraphics[width=0.99\linewidth]
    {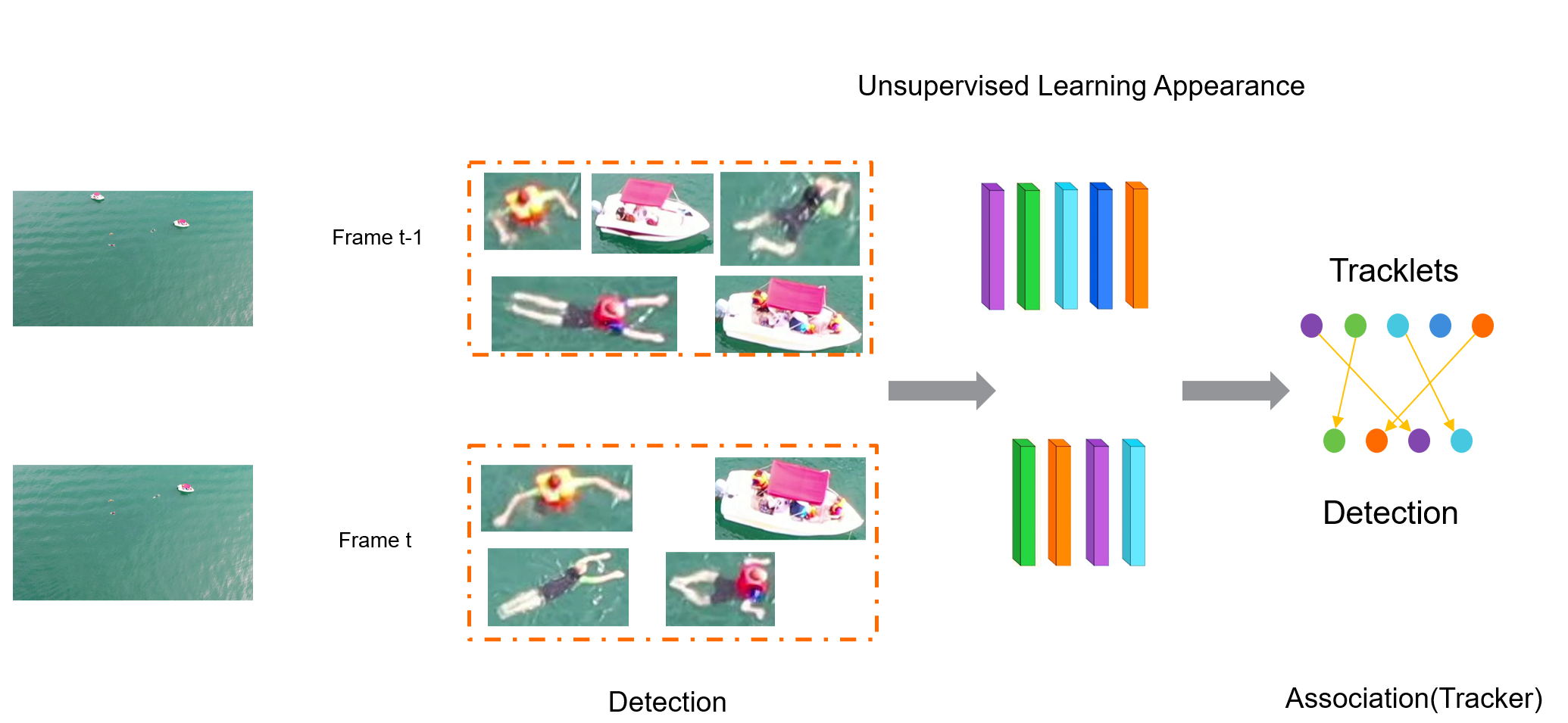}
    \caption{The overall architecture of MOT}
    \label{fig:overview}
\end{figure*}

%-------------------------------------------------------------------------

\section{Method}
In this section, we present detail of the Multiple object tracking framework including \textbf{Overall Architecture}, \textbf{Detection}, \textbf{Appearance Model}, and \textbf{Data Association}.
\subsection{Overall Architecture}

As shown in \textbf{Figure 1}, the proposed method is quite simple and it mainly contains three parts: detection, appearance model(ReID model), and data association(Tracker). The detection part is mainly responsible for providing a high-quality instance box. The appearance model part is mainly responsible for providing high-quality embedding features. The data association(Tracker) part leverages detection output and appearance model to output stable trajectories.

\subsection{Detection}

Due to the high performance of the transformer, we adopt a swin-based transformer\cite{liu2021swin} backbone with Composite Backbone Network V2 (CBNetV2)\cite{liang2021cbnetv2} architecture to predict object bounding box. The CBNetV2 integrates high and low-level features of multiple backbones which connected in parallel. The Feature Pyramid Network (FPN)\cite{fpn2017lin} neck and Hybrid Task Cascade (HTC)\cite{htc2019chen} detector are attached and trained in each backbone as a main branch and an assistant branch. Only the main branch is used in the inference process.
\par
 We use more weight to bound box regression than classification in Loss Function for a more compact detection bound box which will benefit appearance model performance.
\par

%\ref{BDD100K MOT val set result}
\begin{table*}[!ht]
    \centering
    \caption{Results on SeaDronesSee Multi-Object Tracking with Reidentification testset}
    \label{BDD100K MOT test set result}
    \begin{tabular}{c c c c c c c c c}
    \hline
      Method & HOTA $\uparrow$ & mMOTA$\uparrow$ & mIDF1$\uparrow$ & MOTP$\uparrow$ & FP $\downarrow$ &FN $\downarrow$  & IDs  $\downarrow$ & FPS(A100) $\uparrow$ \\
    \hline

\textbf{ReIDTracker\_Sea}        & 0.624 & \textbf{0.781} & 0.713 & 0.204 &  9595 & 11166 & 178 & 3 \\

\hline
    \end{tabular}
\end{table*}

\begin{table*}[!ht]

    \centering
    \caption{Results on BoaTrack}
    \label{Waymo 2D Tracking test set result}
    \begin{tabular}{c c c c c c c c c}
    \hline
      Method & HOTA $\uparrow$ & mMOTA$\uparrow$ & mIDF1$\uparrow$ & MOTP$\uparrow$ & FP $\downarrow$ &FN $\downarrow$  & IDs  $\downarrow$  & FPS(A100) $\uparrow$ \\
    \hline
\textbf{ReIDTracker\_Sea} & \textbf{0.214} &	\textbf{0.105} & \textbf{0.232} & 0.214 & 14476 & \textbf{76651} & 1404 & 3 \\
    \hline
    \end{tabular}
\end{table*}

\subsection{Appearance Model}

 Compared to other methods, we use unsupervised appearance models to address the high cost of video trajectory annotation. Our base appearance model for this framework is MoCo-v2 \cite{he2020momentum} with ResNet50 backbone. The model extracts feature representations from detected boxes. MoCo-v2 model training by imagenet 1K dataset and then fine-tuning on  MOT dataset. We also compare with model training by other contrastive learning methods (SimCLR \cite{chen2020simple}, SimCLRv2 \cite{chen2020simple}, MoCo-v2 \cite{he2020momentum}, etc). We also make a comparison between supervised learning and self-supervised learning. Finally, we draw the conclusion that MoCo-v2 \cite{he2020momentum} has better generalization capacity in the Maritime dataset. Because the last convolution module of resnet50 is more related to classification type, not the general features we want, so we finally removed the network in the final integration (\textbf{Figure 2}).
 
\begin{figure}[h!]
    \centering
    \includegraphics[width=0.99\linewidth]
    {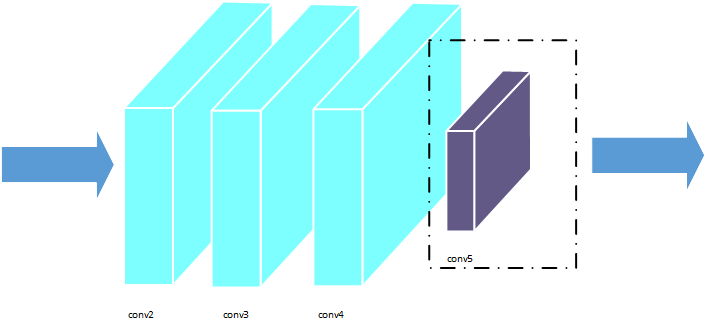}
    \caption{The network of Appearance model}
    \label{fig:overview}
\end{figure}

\subsection{Data Association}
 We adopt the Bytetrack \cite{zhang2021bytetrack} concept which is a simple but strong method for matching object id across frames. The detected boxes in each frame are grouped based on their detection score into the high score and low score. Firstly, the method finds the association between the high score box and the tracklet. Then, the rest of the high score and low score boxes are used to find the association from the remained tracklet. The association method can be different in each association step. 
 \par
 Our method uses only the appearance feature to associate both high and low score boxes with tracklet. In addition, we add a weighted score to tracklet to keep the tracklet representation from the higher detection score since the detection score tends to get lower when the occluded part gets bigger. 
\par
The tracklet features are weighted by the detection score and combined within $\tau$ frames to maintain the object representation during occlusion. The weighted feature $\hat{e_j}$ combined tracklet feature $e_j$ which is weighted by the detection score $s_j$ from the previous $\tau$ frames. 
\par
\begin{equation}
    \hat{e_j}=\frac{\sum^\tau_{t=1}e^t_j\times s^t_j}{\sum^\tau_{t=1}s^t_j}
\end{equation}
\par
$\hat{e_j}$ is further used for finding the matched box in the data association. We apply the same association method with \cite{wang2021different}. A ReId similarity matrix between tracklet and detection box is computed and used to find matching pairs by the Hungarian algorithm \cite{Kuhn1955TheHM}.

\subsection{Implementation Details}
\textbf{Detector}. The Swin-B backbone \cite{liu2021swin} was initiated by a model pre-trained on ImageNet-22K \cite{Russakovsky2014ImageNetLS}. CBNetV2 \cite{liang2021cbnetv2} was trained on the SeaDronesSee Multi-Object Tracking training and validation dataset. We applied multi-scale augmentation to scale the shortest side of images to between 640 and 1280 pixels and applied random flip augmentation during training. Adam optimizer was set with an initial learning rate of 1e-6 and weight decay of 0.05. We trained the model on 4 A100 GPUs with 1 image per GPU for 10 epochs. During inference, we resize an image to 2880x1920 to better detect the small objects. For the detection task, we use a
combination of classification Cross-Entropy loss and the generalized IoU regression loss\cite{rezatofighi2019generalized}. 
Loss weights $\lambda _1$ and  $\lambda _2$   are set to 1.0 and 10.0 by default, which drives the model output more compact bound box.

\par
\begin{equation}
    \mathcal {L} =\lambda _1 \mathcal {L} _{cls}+\lambda _2 \mathcal {L} _{box}
\end{equation}

\indent\textbf{Appearance Model}. The backbone of the appearance model is pre-trained on ImageNet-1K. Then, we fine-tuned the backbone by using MoCo-v2 \cite{he2020momentum} on the SeaDronesSee Multi-Object Tracking dataset. The training dataset contains cropped object images according to bounding box labels from MOT dataset. The optimizer is SGD with a weight decay of 1e-4, a momentum factor of 0.9, and an initial learning rate of 0.12. We trained the model on 4 A100 GPUs with 256 images per GPU.

\indent\textbf{Tracker}. Our method is generally similar to ByteTrack \cite{zhang2021bytetrack}, but we used ReID to match high and low detection boxes.  We set the high detection score threshold to 0.84 and the low detection score threshold to 0.3.
\par

\subsection{Training Data}
for both challenges (SeaDroneSee-MOT and BoaTrack), We use the sample ReID module which training on ImageNet 1K.
\par
\textbf{SeaDronesSee-MOT with Reidentification}:
We just train the detector using the SeaDronesSee MOT trainset and validation set. but we grouped "swimmer" and "swimmer with life jacket" and "life jacket" as one Class.

\par
\textbf{BoaTrack}: we just train the detector using LaRS trainset "boat" annotations.

\subsection{Final Result}

 as shown in Table \ref{BDD100K MOT test set result} and  Table \ref{Waymo 2D Tracking test set result}

\par

%%%%%%%%% REFERENCES

{\small
\bibliographystyle{unsrt}
\bibliography{egbib}

\begin{thebibliography}{10}

\bibitem{liang2021cbnetv2}
Tingting Liang, Xiaojie Chu, Yudong Liu, Yongtao Wang, Zhi Tang, Wei Chu, Jingdong Chen, and Haibin Ling.
\newblock Cbnetv2: A composite backbone network architecture for object detection.
\newblock {\em arXiv preprint arXiv:2107.00420}, 2021.

\bibitem{huang2023reidtrack}
Kaer Huang, Bingchuan Sun, Feng Chen, Tao Zhang, Jun Xie, Jian Li, Christopher~Walter Twombly, and Zhepeng Wang.
\newblock Reidtrack: Multi-object track and segmentation without motion.
\newblock {\em arXiv preprint arXiv:2308.01622}, 2023.

\bibitem{huang1st}
Kaer Huang, Kanokphan Lertniphonphan, Feng Chen, Tao Zhang, Jun Xie, Huabing Liu, Qigang Wang, and Zhepeng Wang.
\newblock 1st place solution for eccv2022 sslad bdd100k mot/mots/ssmot/ssmots challenges.

\bibitem{he2020momentum}
Kaiming He, Haoqi Fan, Yuxin Wu, Saining Xie, and Ross Girshick.
\newblock Momentum contrast for unsupervised visual representation learning.
\newblock In {\em Proceedings of the IEEE/CVF conference on computer vision and pattern recognition}, pages 9729--9738, 2020.

\bibitem{zhang2021bytetrack}
Yifu Zhang, Peize Sun, Yi~Jiang, Dongdong Yu, Zehuan Yuan, Ping Luo, Wenyu Liu, and Xinggang Wang.
\newblock Bytetrack: Multi-object tracking by associating every detection box.
\newblock {\em arXiv preprint arXiv:2110.06864}, 2021.

\bibitem{kiefer20231st}
Benjamin Kiefer, Matej Kristan, Janez Per{\v{s}}, Lojze {\v{Z}}ust, Fabio Poiesi, Fabio Andrade, Alexandre Bernardino, Matthew Dawkins, Jenni Raitoharju, Yitong Quan, et~al.
\newblock 1st workshop on maritime computer vision (macvi) 2023: Challenge results.
\newblock In {\em Proceedings of the IEEE/CVF Winter Conference on Applications of Computer Vision}, pages 265--302, 2023.

\bibitem{kristan2023first}
Matej Kristan, Ji{\v{r}}{\'\i} Matas, Martin Danelljan, Michael Felsberg, Hyung~Jin Chang, Luka~{\v{C}}ehovin Zajc, Alan Luke{\v{z}}i{\v{c}}, Ondrej Drbohlav, Zhongqun Zhang, Khanh-Tung Tran, et~al.
\newblock The first visual object tracking segmentation vots2023 challenge results.
\newblock In {\em Proceedings of the IEEE/CVF International Conference on Computer Vision}, pages 1796--1818, 2023.

\bibitem{liu2021swin}
Ze~Liu, Yutong Lin, Yue Cao, Han Hu, Yixuan Wei, Zheng Zhang, Stephen Lin, and Baining Guo.
\newblock Swin transformer: Hierarchical vision transformer using shifted windows.
\newblock In {\em 2021 IEEE/CVF International Conference on Computer Vision (ICCV)}, pages 9992--10002, 2021.

\bibitem{fpn2017lin}
Tsung-Yi Lin, Piotr Dollár, Ross Girshick, Kaiming He, Bharath Hariharan, and Serge Belongie.
\newblock Feature pyramid networks for object detection.
\newblock In {\em 2017 IEEE Conference on Computer Vision and Pattern Recognition (CVPR)}, pages 936--944, 2017.

\bibitem{htc2019chen}
Kai Chen, Jiangmiao Pang, Jiaqi Wang, Yu~Xiong, Xiaoxiao Li, Shuyang Sun, Wansen Feng, Ziwei Liu, Jianping Shi, Wanli Ouyang, Chen~Change Loy, and Dahua Lin.
\newblock Hybrid task cascade for instance segmentation.
\newblock In {\em 2019 IEEE/CVF Conference on Computer Vision and Pattern Recognition (CVPR)}, pages 4969--4978, 2019.

\bibitem{chen2020simple}
Ting Chen, Simon Kornblith, Mohammad Norouzi, and Geoffrey Hinton.
\newblock A simple framework for contrastive learning of visual representations.
\newblock In {\em International conference on machine learning}, pages 1597--1607. PMLR, 2020.

\bibitem{wang2021different}
Zhongdao Wang, Hengshuang Zhao, Ya-Li Li, Shengjin Wang, Philip Torr, and Luca Bertinetto.
\newblock Do different tracking tasks require different appearance models?
\newblock {\em Advances in Neural Information Processing Systems}, 34:726--738, 2021.

\bibitem{Kuhn1955TheHM}
Harold~W. Kuhn.
\newblock The hungarian method for the assignment problem.
\newblock {\em Naval Research Logistics (NRL)}, 52, 1955.

\bibitem{Russakovsky2014ImageNetLS}
Olga Russakovsky, Jia Deng, Hao Su, Jonathan Krause, Sanjeev Satheesh, Sean Ma, Zhiheng Huang, Andrej Karpathy, Aditya Khosla, Michael~S. Bernstein, Alexander~C. Berg, and Li~Fei-Fei.
\newblock Imagenet large scale visual recognition challenge.
\newblock {\em International Journal of Computer Vision}, 115:211--252, 2014.

\bibitem{rezatofighi2019generalized}
Hamid Rezatofighi, Nathan Tsoi, JunYoung Gwak, Amir Sadeghian, Ian Reid, and Silvio Savarese.
\newblock Generalized intersection over union: A metric and a loss for bounding box regression.
\newblock In {\em Proceedings of the IEEE/CVF conference on computer vision and pattern recognition}, pages 658--666, 2019.

\end{thebibliography}
}

\end{document}